\documentclass[10pt,twocolumn,letterpaper]{article}

\usepackage{cvpr}
\usepackage{times}
\usepackage{epsfig}
\usepackage{graphicx}
\usepackage{amsmath}
\usepackage{amssymb}

\usepackage{bm}
\usepackage{comment}
\usepackage{subfigure}
\usepackage{multirow}
\usepackage{algorithm}
\usepackage{algorithmic}

\usepackage[breaklinks=true,bookmarks=false]{hyperref}

\cvprfinalcopy 

\def\cvprPaperID{****} 

\begin{document}

\renewcommand{\algorithmicrequire}{\textbf{Input:}} 
\renewcommand{\algorithmicensure}{\textbf{Output:}} 

\renewcommand{\a}{{\bm{a}}}
\renewcommand{\b}{{\bm{b}}}
\renewcommand{\c}{{\bm{c}}}
\renewcommand{\d}{{\bm{d}}}
\newcommand{\e}{{\bm{e}}}
\newcommand{\f}{{\bm{f}}}
\newcommand{\g}{{\bm{g}}}
\newcommand{\h}{{\bm{h}}}
\newcommand{\m}{{\bm{m}}}
\renewcommand{\o}{{\bm{o}}}
\newcommand{\p}{{\bm{p}}}
\newcommand{\s}{{\bm{s}}}
\renewcommand{\u}{{\bm{u}}}
\renewcommand{\v}{{\bm{v}}}
\newcommand{\w}{{\bm{w}}}
\newcommand{\x}{{\bm{x}}}
\newcommand{\y}{{\bm{y}}}
\newcommand{\z}{{\bm{z}}}
\newcommand{\balpha}{{\bm{\alpha}}}
\newcommand{\bmu}{{\bm{\mu}}}
\newcommand{\bsigma}{{\bm{\sigma}}}
\newcommand{\blambda}{{\bm{\lambda}}}
\newcommand{\bgamma}{{\bm{\gamma}}}

\newcommand{\A}{{\mathcal{A}}}
\newcommand{\B}{\mathcal{B}}
\newcommand{\C}{\mathcal{C}}
\newcommand{\D}{\mathcal{D}}
\newcommand{\E}{\mathcal{E}}
\newcommand{\F}{\mathcal{F}}
\newcommand{\G}{\mathcal{G}}
\renewcommand{\H}{\mathcal{H}}
\newcommand{\K}{\mathcal{K}}
\renewcommand{\L}{\mathcal{L}}
\newcommand{\W}{\mathcal{W}}
\newcommand{\X}{\mathcal{X}}
\newcommand{\Y}{\mathcal{Y}}

\renewcommand{\P}{\mathcal{P}}
\newcommand{\Q}{\mathcal{Q}}
\newcommand{\R}{\mathbb{R}}
\renewcommand{\S}{\mathbb{S}}

\newcommand{\T}{\mathcal{T}}
\newcommand{\1}{\mathbbm{1}}
\newcommand{\argmin}{{\mbox{argmin}}}

\title{A Semi-Supervised Assessor of Neural Architectures}

\author{Yehui Tang$^{1,2}$, Yunhe Wang$^{2}$, Yixing Xu$^{2}$, Hanting Chen$^{1,2}$, Boxin Shi$^{3,4}$,\\
	Chao Xu$^{1}$, Chunjing Xu$^{2}$\thanks{Corresponding author.}, Qi Tian$^{2}$, Chang Xu$^{5}$ \\
	\normalsize$^1$ Key Lab of Machine Perception (MOE), Dept. of Machine Intelligence, Peking University.
	\\
	\normalsize$^2$ Noah's Ark Lab, Huawei Technologies. \normalsize$^3$ NELVT, Dept. of CS, Peking University. \normalsize$^4$ Peng Cheng Laboratory. \\
	\normalsize$^5$ School of Computer Science, Faculty of Engineering, University of Sydney.\\
	\small\texttt{\{yhtang,chenhanting,shiboxin\}@pku.edu.cn; xuchao@cis.pku.edu.cn}\\
	\small\texttt{\{yunhe.wang,xuyixing,xuchunjing,tian.qi1\}@huawei.com; c.xu@sydney.edu.au}
}

%

\maketitle

\begin{abstract}
 Neural architecture search (NAS) aims to automatically design deep neural networks of satisfactory performance. Wherein, architecture performance predictor is critical to efficiently value an intermediate neural architecture. But for the training of this predictor, a number of neural architectures and their corresponding real performance often have to be collected. In contrast with classical performance predictor optimized in a fully supervised way, this paper suggests a semi-supervised assessor of neural architectures. We employ an auto-encoder to discover meaningful representations of neural architectures. Taking each neural architecture as an individual instance in the search space, we construct a graph to capture their intrinsic similarities, where both labeled and unlabeled architectures are involved. A graph convolutional neural network is introduced to predict the performance of architectures based on the learned representations and their relation modeled by the graph. Extensive experimental results on the NAS-Benchmark-101 dataset demonstrated that our method is able to make a significant reduction on the required fully trained architectures for finding efficient architectures. 
\end{abstract}

\section{Introduction}
The impressive successes in computer vision tasks, such as image classification~\cite{he2016deep,han2018co,han2019ghostnet}, detection~\cite{cai2018cascade} and segmentation~\cite{zhou2019context}, heavily depends on an effective design the backbone deep neural networks, which are usually over-parameterized for the sake of effectiveness. Instead of resorting to human expert experience, Neural Architecture Search (NAS) framework focuses on an automatic way to select hyper-parameters and design appropriate network architectures.

There have been a large body of works on NAS, and they can be roughly divided into two categories. Combinatorial optimization methods search architectures in a discrete space by generating, evaluating and selecting different architectures, e.g. Evolutionary Algorithm (EA) based methods~\cite{real2019regularized} and Reinforcement Learning (RL) based methods~\cite{zoph2016neural}. The other kind of NAS methods are continuous optimization based, which relax the original search space to a continuous space and gradient-based optimization is usually applied~\cite{liu2018darts,li2019random,bender2018understanding,wu2019fbnet,xie2018snas}.  In NAS, to get the exact performance of an architecture, it often takes hours or even days for a sufficient training process. Reducing the number of training epochs or introducing the weight sharing mechanism could alleviate prohibitive computational cost, but it would result in inaccurate performance estimation for the architectures. Recently, there are studies to collect many network architectures with known real performance on the specific tasks  and train a performance predictor~\cite{deng2017peephole,sun2019surrogate}. This one-off training of the predictor can then be applied to evaluate the performance of intermediate searched architectures in NAS, and the overall evaluation cost of an individual architecture can be reduced from hours to milliseconds.  

A major bottleneck in obtaining a satisfactory architecture performance predictor could be the collection of a large annotated training set. Given the expensive cost on annotating a neural architecture with its real performance, the training set for the performance predictor is often small, which would lead to an undesirable over-fitting result. Existing methods insist on the fully supervised way to train the performance predictor, but neglect the significance of those neural architectures without annotations. In the search space of NAS, a number of valid neural architectures can be sampled with ease. Though the real performance could be unknown, their architecture similarity with those annotated architectures would convey invaluable information to optimize the performance predictor.   

In this paper, we propose to assess neural architectures in a semi-supervised way for training the architecture predictor using the well-trained networks as fewer as possible.  Specifically, a very small proportion of architectures are randomly selected and trained on the target dataset to obtain the ground-truth labels. With the help of massive unlabeled architectures, an auto-encoder is used to discover meaningful representations. Then we construct a relation graph involving both labeled and unlabeled architectures to capture intrinsic similarities between architectures. The GCN assessor takes the learned representations of all these architectures  and the relation graph as input to predict the performance of unlabeled architectures. The entire system containing the auto-encoder and GCN assessor can be trained in an end-to-end manner. Extensive experiments results on the NAS-bench-101 dataset \cite{ying2019bench} demonstrate the superiority of the proposed semi-supervised assessor for searching efficient neural architectures.

This paper is organized as follows: in Sec.~2 we briefly review several performance predictors and analyze pros and cons of them, and give an introduction of NAS, GCN and auto-encoder. Sec.~3 gives a detailed implementation of the proposed method. Several experiments conducted on NASBench dataset and the results are shown in Sec.~4. Finally, Sec.~5 summarizes the conclusions.

\section{Related Works}
In this section, we first review current methods of NAS and performance predictor, and then introduce the classical GCN and auto-encoder.
\subsection{Neural Architecture Search (NAS)} 
Current NAS framework for obtaining desired DNNs can be divided into two sub-problems, \ie, search space and search method.

A well-defined search space is extremely important for NAS, and there are mainly three kinds of search spaces in the state-of-the-art NAS methods. The first is cell based search space~\cite{pham2018efficient,zoph2016neural,zoph2018learning,liu2018progressive}. Once a cell structure is searched, it is used in all the layers across the network by stacking multiple cells. Each cell contains several blocks, and each of the block contains two branches, with each branch applying an operation to the output of one of the former blocks. The outputs of the two branches are added to get the final output of the block. The second is Direct Acyclic Graph (DAG) based search space~\cite{ying2019bench}. The difference between cell based and DAG based search space is that the latter does not restrict the number of branches. The input and output number of a node in the cell is not limited. The third is factorized hierarchical search space~\cite{tan2019mnasnet,wu2019fbnet,guo2019single}, which allows different layer architectures in different blocks.

Besides search space, most of the NAS research focus on developing efficient search methods, which can be divided into combinatorial optimization methods and continuous optimization methods\cite{liu2017hierarchical,xue2019transferable,xu2019rnas,liu2018darts}. Combinatorial optimization methods include Evolutionary Algorithm (EA) based methods~\cite{liu2017hierarchical,miikkulainen2019evolving,real2019regularized,real2017large,yang2019cars} and Reinforcement Learning (RL) based methods~\cite{zoph2016neural,zoph2018learning,baker2016designing}. Continuous optimization methods include DARTS~\cite{liu2018darts}, which makes the search space continuous by relaxing the categorical choice of a particular operation to a softmax over all possible operations, and several one-shot methods that solve the problem in a one-shot procedure~\cite{pham2018efficient}. Recently, architecture datasets with substantial full-trained neural architectures are also  proposed to compare different NAS methods conveniently and fairly~\cite{ying2019bench,dong2020bench,zela2020bench}.

\subsection{NAS Predictor}
There are limited works focusing on predicting the network performance. Some of the previous works were designed on hyper-parameter optimization with Gaussian Process~\cite{swersky2014freeze}, which focus on developing optimization functions to better evaluate the hyper-parameter. Other methods directly predict the performance of a given network architecture. The first way is to predict the final accuracy by using part of the learning curves with a mixture of parametric functions~\cite{domhan2015speeding}, Bayesian Neural Network~\cite{klein2016learning} or $v$-SVR~\cite{baker2017practical}. The second way is to predict the performance of a network with a predictor. Deng \etal~\cite{deng2017peephole} extract the feature of a given network architecture layer by layer, and the features with flexible length are sent to LSTM to predict the final accuracy. Istrate \etal~\cite{istrate2019tapas} use a similar manner to predict the accuracy with random forest, believing that few training data are required by using random forest. Luo \etal~\cite{luo2018neural} propose an end-to-end manner by using an encoder to extract features of the networks. The learned features are optimized with gradient descent and then decoded into new architectures with an decoder. The architecture derived in this way is regarded as the optimal architecture with a high performance.
\begin{figure*}[t] 
	\centering
	\includegraphics[width=1.0\linewidth]{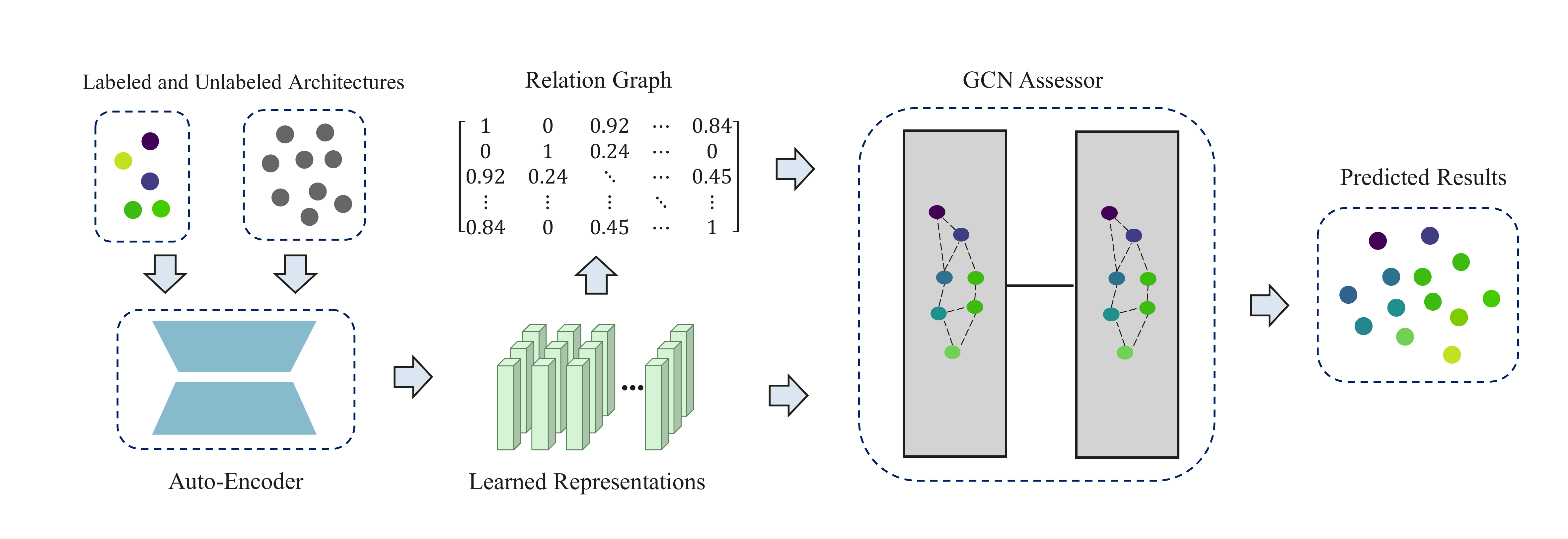}
	\caption{Performance prediction pipeline of the proposed semi-supervised assessor. Both labeled and unlabeled architectures are sent to the auto-encoder to get the meaningful representations. Then a relation graph is constructed to capture architecture similarities based the learned representations. Both the representations and relation graph are sent to the GCN assessor to outputs estimated performance of architectures. The entire system can be trained end-to-end.}
	\label{fig-pipeline}
\end{figure*}
\subsection{Graph Convolutional Network (GCN)} 
GCN is a prevalent technique tackling data generated from non-Euclidean domains and represented as graphs with complex relation. Sperduti \etal \cite{sperduti1997supervised} first tackle DAGs with neural networks and recently GCNs achieve the-state-of-art performance in multiple tasks, such as citation networks \cite{kipf2016semi}, social networks \cite{li2019semi} and point clouds data analyses \cite{zhang2018graph}. Both graph-level task and node level tasks can be tackled with GCNs. For a graph-level task, each graph is seen as an individual and the GCN is to predict the labels of those graphs. As for node-level tasks, the examples are seen as vertices of a graph which reflects the relation between them,  and the labels of examples are predicted by the GCN with the help of graph. Beyond the features of examples, the graph also provides extra valuable information and improves prediction accuracy.

\section{Approach}

Consider the search space $X= X^l \bigcup X^u$ with $N=N_l+N_u$ architectures, where $X^l=\{\x^l_{1},\x^l_{2}, \cdots, \x^l_{N_l}\}$ are annotated architectures with the corresponding ground-truth performance $\y^l=\{y^l_1, y^l_2, \cdots, y^l_{N_l} \}$, and $X^u=\{\x^u_{1},\x^u_{2}, \cdots, \x^u_{N_u}\}$ are the remaining massive unlabeled architectures.  The assessor $\P$ is to take the architecture $x_i\in X$ as the input and output the estimated performance $\hat y_i =\P(W_p,\x_i)$, where $W_p$ is the trainable parameters of the assessor $\P$. Given a sufficiently large labeled architecture set as the training data, the assessor $\P$ can be trained in a supervised manner to  fit the ground truth performance \cite{deng2017peephole,sun2019surrogate}, \ie, 
\begin{equation}
	\min_{W_p} \frac{1}{N_l} \sum_{i=1}^{N_l} ||\P(W_p,\x^l_i) -y^l_i||^2_2,
\end{equation}
where $||\cdot||_2$ denotes $\ell_2$ norm. However, due to the limitation of time and computational resources, very limited architectures can be trained from scratch to get the ground-truth performance, which would not be enough to support the training of a predictor with high accuracy. Actually, there are  massive architectures without annotations and they can participate in the prediction progress. The similarity between architectures can provide extra information to make up the insufficiency of labeled architectures and help training the performance predictor to achieve higher performance.   

\subsection{Architecture Embedding}
\label{sc-graph}

Before sending neural architectures to the performance predictor, we need an encoder $\E$ is to get the appropriate embedding of architectures. There are already some common  hand-crafted representations of architectures for specific search spaces. For example, Ying~\etal~\cite{ying2019bench}  represent  the architectures in a  Directed Acyclic Graph (DAG) based search space with adjacency matrices,  where 0 represents no connection between two nodes and the non-zero integers denote the operation types. Though these hand-crafted representations can describe different architectures, they are usually redundant and noisy to express the intrinsic property of architectures. In contrast with this manual approach, we aim to discover more effective representations of neural architectures with an auto-encoder.  

A classical auto-encoder \cite{khanum2015survey} contains two modules: the encoder $\E$ and decoder $\D$. $\E$ takes the hand-crafted representations of both labeled architectures $\x^l \in X^l$ and unlabeled architectures $\x^u \in X^u$ as input and maps them to a low-dimension space. Then the learn compact representation are sent to the decoder $\D$ to reconstruct the original input. The auto-encoder is trained as:
\begin{align}
	\label{rc}
	\begin{split}
		\min_{W_e,W_d} \L_{rc}& =  \frac{1}{N_l} \sum_{i=1}^{N_l} || \D(\E( \x^l_i;W_e);W_d)- \x^l_i||^2_2\\
		&+ \frac{1}{N_u} \sum_{j=1}^{N_u} || \D(\E( \x^u_j;W_e);W_d)- \x^u_j||^2_2,
	\end{split}
\end{align}
where $W_e$ and $W_d$ are the trainable parameters of the encoder $\E$ and decoder $\D$, respectively\footnote{$\x_i^l$ and $\x_i^u$ in Eq.~(\ref{rc}) also denote the  hand-crafted representations of the architectures without ambiguity.}. The feature $\E(\x_i)$ for architectures $\x_i \in X$ learned by the auto-encoder can be more compact representations of architectures. Most importantly, the auto-encoder can be optimized together with the predictor $\P$ in an end-to-end manner, which enables feature $\E(\x_i)$ to be more compatible with $\P$ to predict the performance of architectures.

\subsection{Semi-supervised Architecture Assessor}
The architectures in a search space are not independent and there are some intrinsic relation between architectures.  For example, an architecture can always be obtained by slightly modifying a very `similar' architecture, such as replacing an operation type, adding/removing an edge, changing the width/depth and so on. Most importantly,  beyond the limited labeled architectures, the massive unlabeled architectures in search space  would also be helpful for the training of assessor $\P$, because of their underlying connections with those labeled architectures. Though obtaining the real performance of all architectures is impossible, exploiting the large volume of unlabeled architectures and exploring intrinsic constraints underlying different architectures will make up the insufficiency of labeled architectures.    

Based on the learned representation $\E(\x_i)$ of architectures, we adopt the common Radial Basis Function (RBF)~\cite{good1993rapid}  to define the similarity measure $s(\x_i,\x_j)$ between architectures $\x_i \in X$ and $\x_j \in X$ , \ie, 
\begin{equation}
	\label{similarity}
	s(\x_i,\x_j)= \exp \left (- \frac{d(\E(\x_i),\E(\x_j))}{2\sigma^2}\right),
\end{equation}
where $d(\cdot,\cdot)$ denotes the  distance measure (\eg, Euclidean distance) and $\sigma$ is a scale factor. $s(\x_i,\x_j)$ ranges in [0,1] and $s(\x_i,\x_i)=1$. When the distance between representation $\E(\x_i)$ and $\E(\x_j)$ becomes larger, the similarity $s(\x_i,\x_j)$ decreases rapidly. 

Given this similarity measurement, the relation between architectures can be easily modeled by a graph $G$, where individual vertex denotes an architecture $\x_i\in X$ and the edge reflects the similarity between architectures. Both labeled and unlabeled architectures are involved in the graph $G$. Denote the adjacency matrix of graph $G$ as $A \in \R ^{N \times N}$, where $A_{ij}=s(\x_i,\x_j)$ if $s(\x_i,\x_j)$ exceeds the threshold $\tau$ and zero otherwise. Note that $A_{ii}=1$ and there are  self-connections in graph $G$. Two similar architectures thus tend to locate close with each other in the graph and are connected by edges with a large weight. The architectures connected by edges have direct relation while those disconnected architectures interact with each other in an implicit way via other vertices. This is accordant to the intuition that two very different architectures can be connected by some intermediate architectures.  

To utilize both limited labeled architectures and  massive unlabeled architectures with their similarity modeled by the graph $G$, we construct the assessor $\P$ by stacking multiple graph convolutional layers\cite{kipf2016semi}, which takes the learned representations of both labeled and unlabeled architectures as inputs. The graph $G$ is also embedded into each layer and guides the information propagation between the features of different architectures. Taking all these architectures as a whole and utilizing the relation between architectures, the assessor $\P$ outputs their estimated performance. A assessor $\P$ composing of two graph convolutional layers is: 
\begin{align}
	\label{gcn}
	\begin{split}
		[\hat \y^l, \hat \y^u ] &=\P(\E([X^l,X^u]),G,W_p) \\
		& =\hat A {\rm ReLU} \left(\hat A \E([X^l,X^u]) W_p^{(0)} \right) W_p^{(1)},
	\end{split}
\end{align}
where $\E([X^l,X^u])$ denotes the learned representations of both labeled and unlabeled architectures, and $\hat \y^l=\{ \hat y^l_1,\hat y^l_2, \cdots, \hat y^l_{N_l}\}$ and $\hat \y^u =\{ \hat y^u_1,\hat y^u_2, \cdots, \hat y^u_{N_u}\}$ are their estimated performance, respectively.  $ D$ is a diagonal matrix where $ D_{ii}= \sum_j {A}_{ij}$, and $\hat A =  D^{-\frac{1}{2}}{A}D^{-\frac{1}{2}}$.  $W^{(0)}_p$, $W^{(1)}_p$ are the weight matrices.

As shown in Eq. (\ref{gcn}), the output of the assessor $\P$ depends on not only their input representation but also the neighboring architectures in the graph $G$ due to adjacency matrix $A$, and thus the performance prediction processes of labeled and unlabeled architectures interact with each other. In fact, GCN can be considered as a Laplacian smoothing operator ~\cite{li2018deeper} and  intuitively, two connected nodes on the graph tend to have similar features and produce similar outputs. As both labeled and unlabeled architectures are sent to the predictor simultaneously, their intermediate features interrelate with each other.

The assessor $\P$ is trained to fit the ground-truth performance of labeled architectures based as both the architectures themselves and the relation between them, \ie, 
\begin{equation}
	\label{lossgcn}
	\min_{W_p} \L_{rg} =\frac{1}{N_l}\sum_{i=1}^{N_l} ||\hat y^l_i - y^l_i ||^2_2,
\end{equation} 
where $W_p$ is the trainable parameter of assessor $\P$. Though the supervised loss is only applied on labeled architectures, the unlabeled architectures also participate in the performance prediction of the labeled architectures via the relation graph $G$, and thus the supervision information from those limited performance labels can guide the feature generation process of those unlabeled architectures. Intuitively, the labels can propagate along the edge in the relation graph $G$, considering the length of paths and the weights of edges. What's more, the training process helps the predictor learn to predict the performance of a given  architecture with the assistance of its neighbors in the graph $G$, which makes the prediction more robust and improve the prediction accuracy.

\subsection{Optimization}
\label{sc-opt}
The auto-encoder and  assessor can constitute an end-to-end system, which learns the representations of architectures and predict performance simultaneously. As shown in Figure \ref{fig-pipeline}, the hand-crafted representations of both labeled architectures $X^l$ and unlabeled architectures $X^u$ are first delivered into the encoder $\E$ to produce learned representations $\E([X^l,X^u])$, and then the relation graph $G$ is constructed based on the representation $\E([X^l,X^u])$ via Eq.~(\ref{similarity}). Both the representation $\E([X^l,X^u])$ and relation graph $G$ are sent to the GCN assessor $\P$ to get the estimated performance $\hat \y$. In the training phase, the learned representations $\E([X^l,X^u])$ are also sent to the decoder $\D$ to reconstruct the original input. Combining the regression loss $\L_{rg}$ that fits the ground-truth performance and the reconstruction loss $\L_{rc}$ , the entire system is trained as: 
\begin{equation}
	\label{all}
	\min_{W_e,W_d, W_p} \L=(1-\lambda)  \L_{rg} +  \lambda \L_{rc}. 
\end{equation}
where $\lambda \in [0,1]$ is the hyper-parameter that balances the two types of loss functions.  In the end to end system, the learning of architecture representations and performance prediction are promoted mutually. The regression loss $\L_{rg}$ focuses  on fitting the ground-truth performance of labeled architectures and  propagating labels to the unlabeled architectures, which also makes the learned representations $\E([X^l,X^u])$ have stronger relativity to the ground-truth performance. The reconstruction loss $\L_{rc}$ refines information from the massive unlabeled architectures to supply the limited labeled examples and makes the training process more robust. Note that for both regression loss $\L_{rg}$ and reconstruction loss $\L_{rc}$, the unlabeled architectures participate in their optimization process and play an important role.

When implementing the proposed semi-supervised assessor to a large search space containing massive architectures, it is inefficient to construct a large graph containing all the $N$ architectures. Constructing the graph needs to calculate the similarity of arbitrary two architectures which is time-consuming, and storing such a graph also needs a large memory. Mini-batch is a common strategy to tackle big data in deep learning~\cite{krizhevsky2012imagenet}, and  we propose to construct the graph and train the entire system with mini-batch. For each mini-batch, labeled and unlabeled architectures are randomly sampled from $X^l$ and $X^u$, and the graph is constructed with those examples. Thus the entire system can be trained  efficiently with random gradient descent on memory-limited GPUs. The mini-batch training algorithm is presented in Algorithm~\ref{alg}. 
\begin{algorithm}[tb]
	\caption{Training of the semi-supervised assessor.} 
	\label{alg}
	\begin{algorithmic}[1]		
		\REQUIRE{Search space $X=X^l \bigcup X^u$, and the ground-truth performance $\y^l$ for labeled architectures.}		
		\REPEAT
		\STATE Randomly select labeled and unlabeled architectures from  $X^l$ and $X^u$ respectively to form a mini-batch $\B$;
		\STATE Send the architectures $\x \in \B$ to feature extractor $\E$ and get the learned representation $\E(\x)$;
		\STATE Calculate the similarity between architectures via Eq.~(\ref{similarity}) and construct the relation graph $G$;
		\STATE Send the learned representation $\E(\x)$ and relation graph $G$ to the GCN assessor $\P$ and output the approximate performance $\hat y$;
		\STATE Calculate the regression loss $\L_{rg}$ via Eq.~(\ref{lossgcn}); 		
		\STATE Send the learned representation $\E(\x)$ to  the  decoder $\D$ and  calculate the reconstruction loss $\L_{rc}$ via Eq.~(\ref{rc});
		\STATE Calculate the final loss $\L=(1-\lambda)  \L_{rg} +  \lambda \L_{rc}$;
		\STATE Backward and update the parameters of encoder $\E$, assessor $\P$ and decoder $\D$;
		\UNTIL Convergence;
		\ENSURE{The trained encoder $\E$ and assessor $\P$.}
	\end{algorithmic}
\end{algorithm}

\begin{table*}
	\caption{Comparison of performance prediction results on Nas-Bench-101 dataset.}
	\label{tb-arc}
	\centering
	\begin{tabular}{c||c|c|c|c}
		\hline
		$N_l$ &Criteria&Peephole~\cite{deng2017peephole}&E2EPP~\cite{sun2019surrogate}& Ours\\ \hline \hline
		\multirow{3}{*}{1k}&KTau&$0.4373_{\pm0.0112}$&$0.5705_{\pm0.0082}$&$\bm{0.6541}_{\pm0.0078}$\\
		&MSE&$0.0071_{\pm0.0005}$&$0.0042_{\pm0.0003}$&$0.0031_{\pm0.0003}$\\
		&r&$0.4013_{\pm0.0092}$&$0.4467_{\pm0.0071}$&$0.5240_{\pm0.0068}$\\ \hline
		\multirow{3}{*}{10k}&KTau&$0.4870_{\pm0.0096}$&$0.6941_{\pm0.0058}$&$\bm{0.7814}_{\pm0.0042}$\\
		&MSE&$0.0037_{\pm0.0004}$&$0.0032_{\pm0.0003}$&$0.0026_{\pm0.0002}$\\
		&r&$0.4672_{\pm0.0075}$&$0.6164_{\pm0.0063}$&$0.6812_{\pm0.0051}$\\ \hline
		\multirow{3}{*}{100k}&KTau&$0.4976_{\pm0.0055}$&$0.7004_{\pm0.0051}$&$\bm{0.8456}_{\pm0.0031}$\\
		&MSE&$0.0036_{\pm0.0003}$&$0.0024_{\pm0.0002}$&$0.0016_{\pm0.0002}$\\
		&r&$0.4804_{\pm0.0074}$&$0.5874_{\pm0.0051}$&$0.8047_{\pm0.0049}$\\ \hline
	\end{tabular}	
\end{table*}

\section{Experiments}
In this section, we conduct extensive experiments to validate the effectiveness of the proposed semi-supervised assessor. Firstly, the performance prediction accuracies of our method are compared with several state-of-the-art methods. Then we embed the proposed assessor and peer competitors to the combinatorial searching algorithm (such as evolutionary algorithm) to identify architectures with good performance. Ablation studies are also conducted to further analyze the proposed method. 

\textbf{Dataset.} Nas-Bench-101 \cite{ying2019bench} is the largest public architecture dataset for NAS research proposed recently, containing 423K unique CNN architectures trained on CIFAR-10 \cite{krizhevsky2009learning} for image classification, and the best architecture achieves a test accuracy of 94.23\%. The search space for Nas-Bench-101 is a feed-forward structure stacked by blocks and each block is constructed by stacking the same cell 3 times. As all the network architectures in the search space are trained completely to get their ground-truth performance, it is fair and convenient to compare different performance prediction methods comprehensively on Nas-Bench-101. A more detailed description of  the dataset can be referred to \cite{ying2019bench}. Besides Nas-Bench-101, we also construct a small architecture dataset on CIFAR-100 to verify the effectiveness of the methods on different datasets. 

\textbf{Implementation details.} 
The encoder $\E$ is constructed by stacking two convolutional layers followed by a full-connected layer and the decoder $\D$ is the reverse. The inputs of $\E$ are the matrix representations of architectures following \cite{ying2019bench,xu2019rnas}. The assessor $\P$ consists of two graph convolutional layers and outputs the predicted performance. The scale factor $\sigma$ and threshold $\tau$ for constructing graph are set to 0.01 and $10^{-5}$, and $\lambda$ in Eq.~(\ref{all}) is set to 0.5 empirically. The entire system is trained end-to-end with Adam optimizer \cite{kingma2014adam} without weight decay for 200 epochs\footnote{The auto-encoder is first pre-trained as initialization for optimization stabilization.}. The batch size and initial learning rate are set to 1024 and 0.001, respectively. All the experiments are conducted with Pytorch library\cite{paszke2017automatic} on NVIDIA V100 GPUs.
\subsection {Comparison of Prediction Accuracies}
We compare the proposed method with the state-of-the-art predictors based methods  Peephole \cite{deng2017peephole} and E2EPP~\cite{sun2019surrogate}. Since the main function of the performance predictors is to identify better architectures in a search space, accurate performance ranking of architectures is more important than their absolute values. KTau~$\in[-1,1]$ is a common indicator measuring the correlation between the ranking of prediction values and the actual labels, and higher values mean more accurate prediction. Two other common criteria mean square error (MSE) and correlation coefficient (r) are also compared for completeness. MSE measures the deviation of predictions from the ground truth directly, and r~$\in[-1,1]$ measures the correlation degree between prediction values and true labels.

The experimental results are shown in Table \ref{tb-arc}. We randomly sample $N_l$ architectures from the search space (including 423k architectures) as labeled examples, and varies $N_l$ from \{1k, 10k, 100k\}. All possible architectures
are available once the search space has been given, and thus the remaining architectures are used as unlabeled architectures, \ie, $N^u=N-N^l$. As shown in Table \ref{tb-arc}, the proposed semi-supervised assessor surpasses the state-of-the-art methods on three criteria with different number of labeled examples. For example, with 1k labeled architectures, KTau of our method can achieve 0.6541, which is 0.2168 higher than Peephole~(0.4373) and 0.0836 higher than E2EPP~(0.5705), meaning more accurate predicted ranking. The correlation coefficient $r$ is also improved by 0.1227 and 0.0773, indicating higher correction between predicted values and ground-truth labels using our method. The improved performance comes from more thorough exploitation of the information in the search space, which makes up the insufficiency of labeled data. Note that increasing $N_l$ improves the performance of all these methods, but the computational cost of training these architectures is also increased. Thus, the balance between the performance of the predictors and the computation cost of getting labeled examples  needs to be considered in practice.

\begin{figure}[htb] 
	\subfigure[Peephole]{
		\begin{minipage}[t]{0.3\linewidth}
			\centering
			\includegraphics[width=0.99\linewidth]{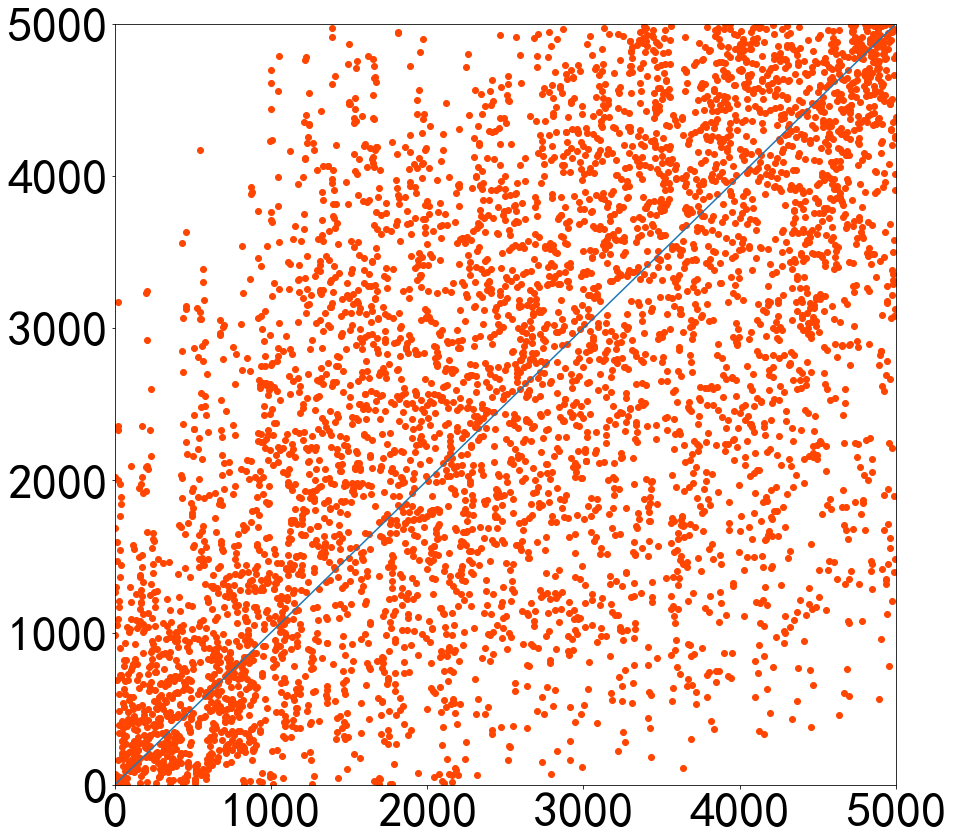}
		\end{minipage}
	}
	\subfigure[E2EPP]{
		\begin{minipage}[t]{0.3\linewidth}
			\centering
			\includegraphics[width=0.99\linewidth]{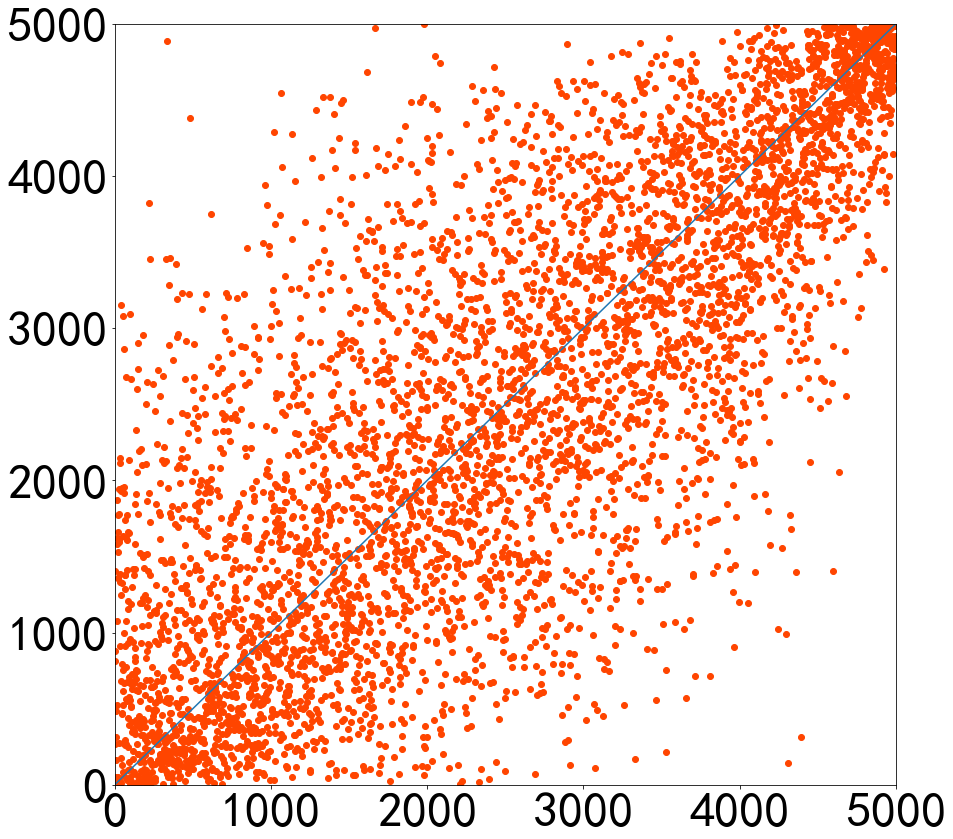}
		\end{minipage}
	}
	\subfigure[Ours]{
		\begin{minipage}[t]{0.3\linewidth}
			\centering
			\includegraphics[width=0.99\linewidth]{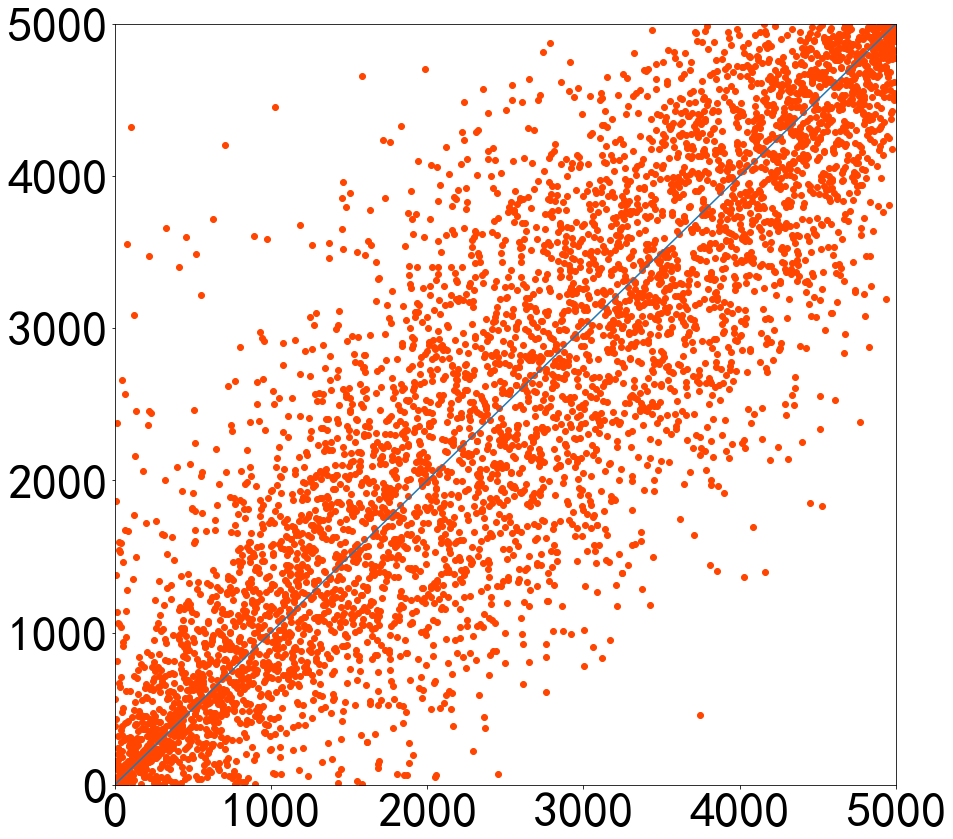}
		\end{minipage}
	}	
	\caption{Predicted ranking of architectures and the corresponding true ranking on Nas-Bench-101 dataset. The $x$-axis denotes the true ranking and $y$-axis denotes the predicted ranking.}
	\label{fig-qual}
	
\end{figure}

The qualitative results are shown in Figure~(\ref{fig-qual}). For clarity, 5k architectures are randomly sampled and shown in the scatter diagrams. The $x$-axis of each point (architecture) is its  ground truth ranking and the $y$-axis is predicted ranking. For our method the points  are much closer to the diagonal line, implying stronger consistency between the predicted ranking and ground truth ranking. Both the numerical criteria and intuitive diagrams show that our method surpasses the state-of-the-art methods. 

\subsection{Searching Results on NAS-Bench-101}

The performance predictors  can be embedded to various architecture search algorithms \cite{sun2019surrogate}such as random search, Reinforcement learning (RL) based methods~\cite{zoph2016neural} and Evolutionary Algorithm  (EA) based methods~\cite{real2019regularized}. Taking EA based methods as an example, the performance predicted by the predictors can be used as fitness, and other progresses including population generation, cross-over and mutation are not changed.  Since we focus on the design of performance predictors, we embed different prediction methods into EA to find the architectures with high performance.  Concretely, we compare the best performance among the top-10 architectures selected by different methods, and all the methods are repeated 20 times with different random seeds.

The performance of the  best architecture selected by different methods is shown in Table \ref{tb-cifar10}. The second column is the accuracies of architectures on CIFAR-10 dataset and the third column is their real performance rankings in all the architectures of Nas-Bench-101. The best network identified by the proposed semi-supervised assessor achieves performance 94.01\%, outperforming the compared methods (93.41\% for Peephole and 93.77\%) by a large margin, since the proposed method can make a more accurate estimation of performance and further identify those architectures with better performance. Though only 1k architectures are sampled to train the predictor, it can still find the architectures whose real performance is in the top 0.01\% of the search space. Compared to the global best architecture with performance 94.23\%, which is obtained by exhaustively enumerating all the possible architectures in the search space, the performance 94.01\% obtained by our method with only 1k labeled architectures is comparable.

\begin{figure}[t] 
	\centering
	\includegraphics[width=0.65\columnwidth]{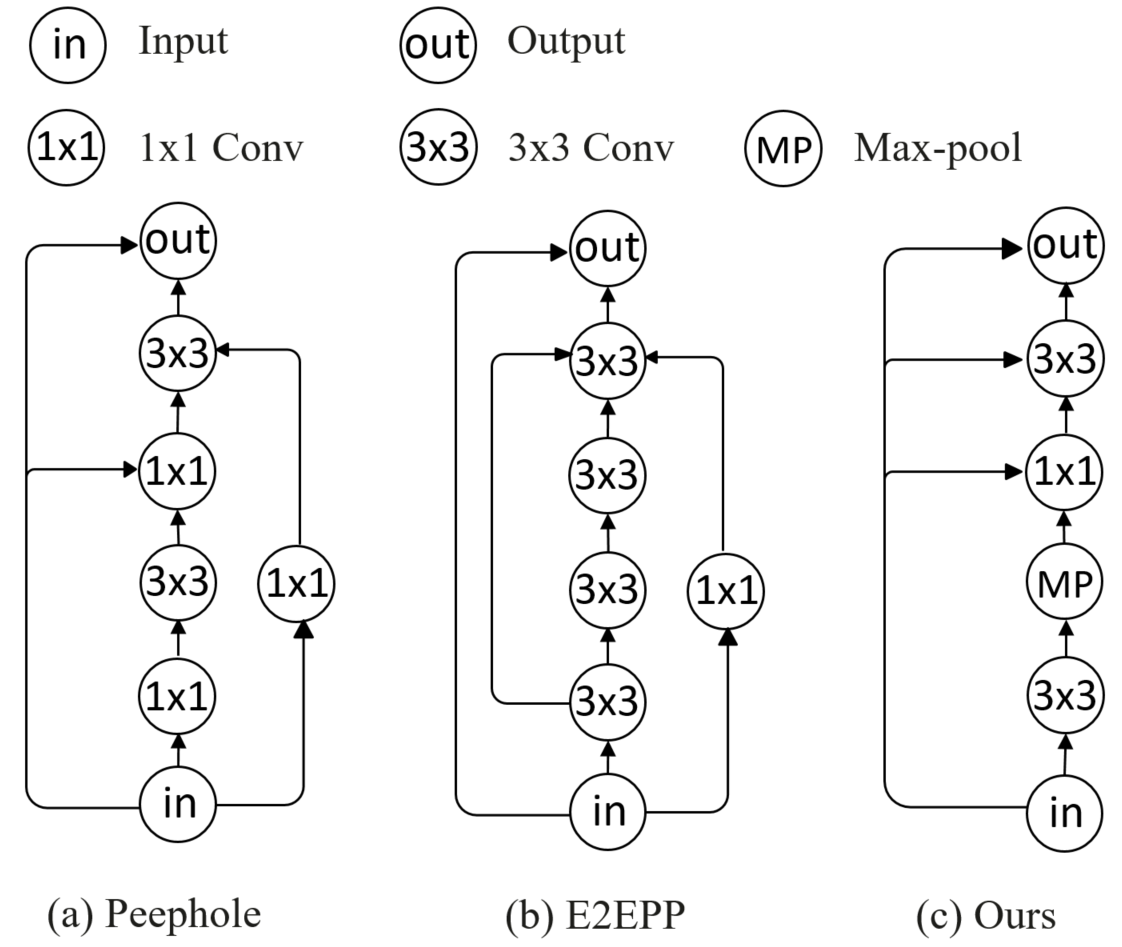}%

	\caption{Visualization of the best network architectures selected by different methods. 1k architectures randomly selected from Nas-bench-101 are used as annotated examples.}
	\label{fig-arc}
\end{figure}

We further show the intuitive representation of the best architectures identified by different methods in Figure \ref{fig-arc}. There are some common characteristics for these architectures with good performance, \eg, both existing a very short path (\eg, path length 1) and a long path  from the first node to the last. The long path consisting of multiple operations ensures the representation ability of the networks, and the short path makes gradient propagate easily to the shallow layers.
The architecture identified by our method (Figure~\ref{fig-arc}(c)) also contains a max pooling layer in the longest path to enlarge the receptive field, which may be a reason for the better performance.    

\begin{table}[t]
	\caption{Classification accuracies on CIFAR-10 and the performance ranking among all the architectures of Nas-Bench-101. 1k architectures randomly selected from Nas-Bench-101 are used as annotated examples.}
	\label{tb-cifar10}
	\centering
	\small
	\begin{tabular}{l|c|c}
		\hline
		Method & Top-1 Accuracy (\%) & Ranking (\%) \\ \hline \hline
		Peephole~\cite{deng2017peephole} &93.41$\pm$0.34 & 1.64\\ 
		E2EPP~\cite{sun2019surrogate} & 93.77$\pm$0.13& 0.15 \\ \hline 
		Ours &\textbf{94.01}$\pm$0.12  & \textbf{0.01} \\ 
		\hline
	\end{tabular}	
\end{table}

\subsection{Experiments on CIFAR-100 Dataset}
To verify the effectiveness of the proposed semi-supervised assessor in different datasets, we further conduct experiments on the common object classification dataset CIFAR-100. Since there is no architecture dataset with ground-truth performance  based on CIFAR-100, we randomly sample 1k architectures from the search space of NAS-bench-101 and train them completely from scratch using the same training hyper-parameters in \cite{ying2019bench}. With the 1k labeled architectures, different performance prediction methods are embedded into the EA to find the best performance.  As CIFAR-100 contains 100 categories, we both compare the top-1 and top-5 accuracies. The best performance among the top-10 architectures are compared and all the  methods are repeated for 20 times with different random seeds. 
\begin{table}[t]
	\caption{Classification accuracies of the best network architectures on CIFAR-100 selected by different methods. 1k network architectures trained on CIFAR-100 are used as annotated examples.}
	\label{tb-cifar100}
	\centering
	\small
	\begin{tabular}{l|c|c}
		\hline
		Method & Top-1 Accuracy (\%)& Top-5 Accuracy (\%) \\ \hline \hline
		Peephole~\cite{deng2017peephole} &74.21$\pm$0.32 &92.04$\pm$0.15\\
		E2EPP~\cite{sun2019surrogate} &75.86$\pm$0.19 &93.11$\pm$0.10 \\ \hline
		Ours &\textbf{78.64}$\pm$0.16& \textbf{94.23}$\pm$0.08 \\ 
		\hline
	\end{tabular}	
\end{table}

\begin{figure}[t] 
	\centering
	\includegraphics[width=0.7\columnwidth]{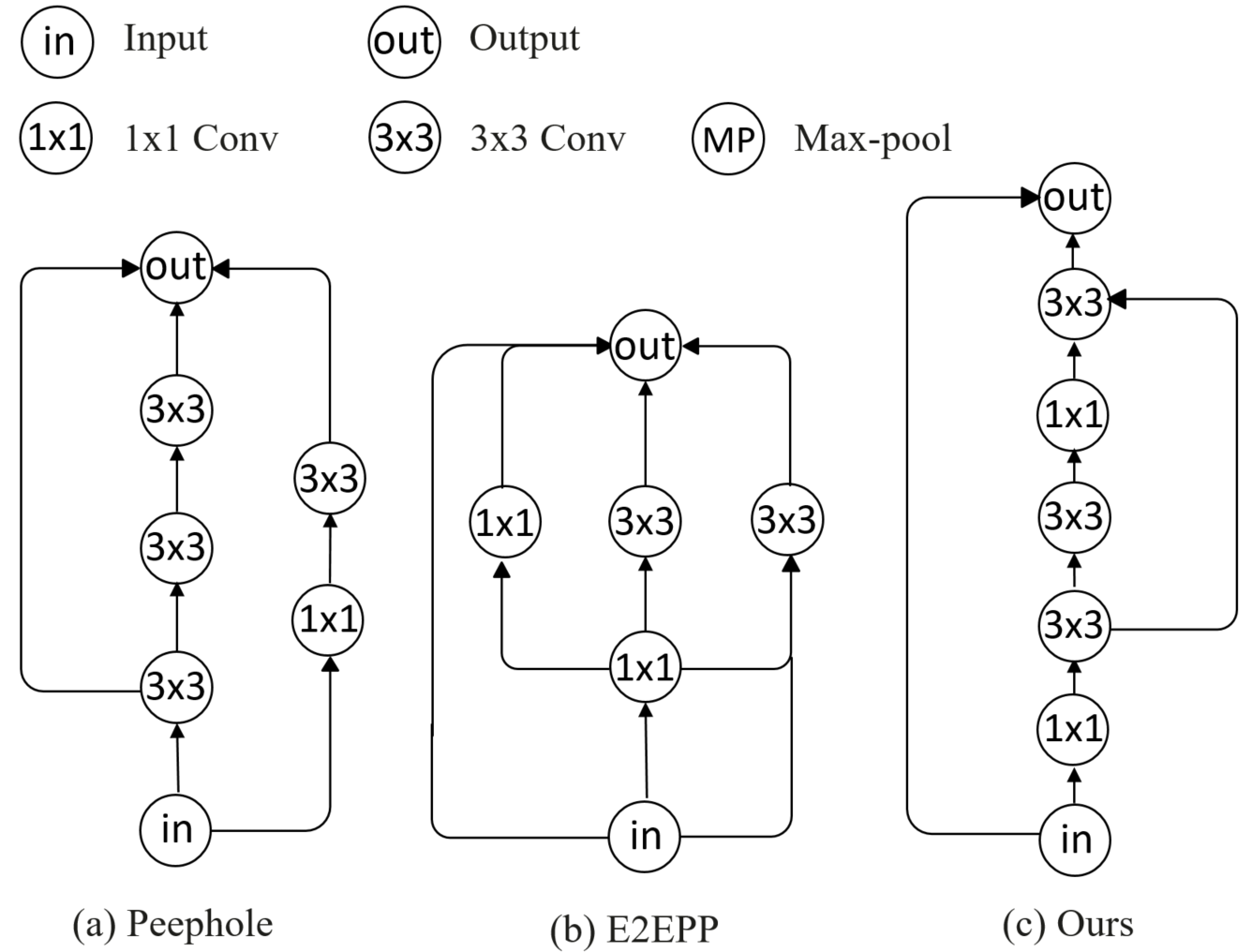}%
	\caption{Visualization of the best network architectures selected by different methods. 1k network architectures trained on CIFAR-100  are used as annotated examples.}
	\label{fig-arc100}
\end{figure}

The accuracies and diagrams of different architectures are shown in Table \ref{tb-cifar100} and Figure \ref{fig-arc100}, respectively. The best architecture identified by our method achieves much higher performance (78.64\% for top-1 and 94.23\% for top-5) compared with the state-of-the-art methods (\eg, 75.86\% for top-1 and 93.11\% for top-5 in E2EPP). It implies that exploring the  relation between architectures and utilizing the massive unlabeled examples in the proposed method works well in different datasets.  
\begin{figure*}[htb] 
	\subfigure[$\sigma$]{
		\begin{minipage}[t]{0.33\linewidth}
			\centering
			\includegraphics[width=0.99\linewidth]{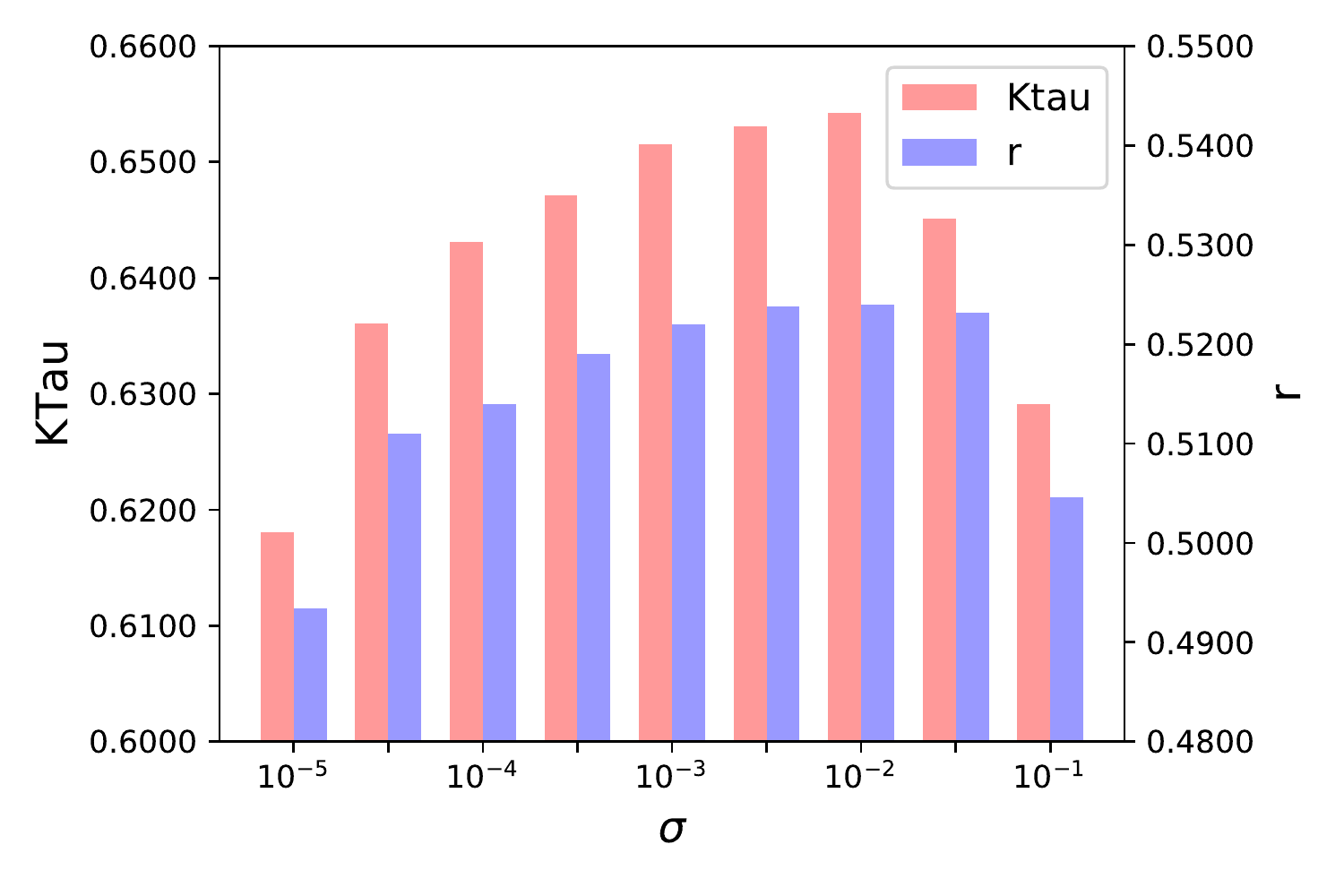}
		\end{minipage}
	}
	\subfigure[$\lambda$]{
		\begin{minipage}[t]{0.33\linewidth}
			\centering
			\includegraphics[width=0.99\linewidth]{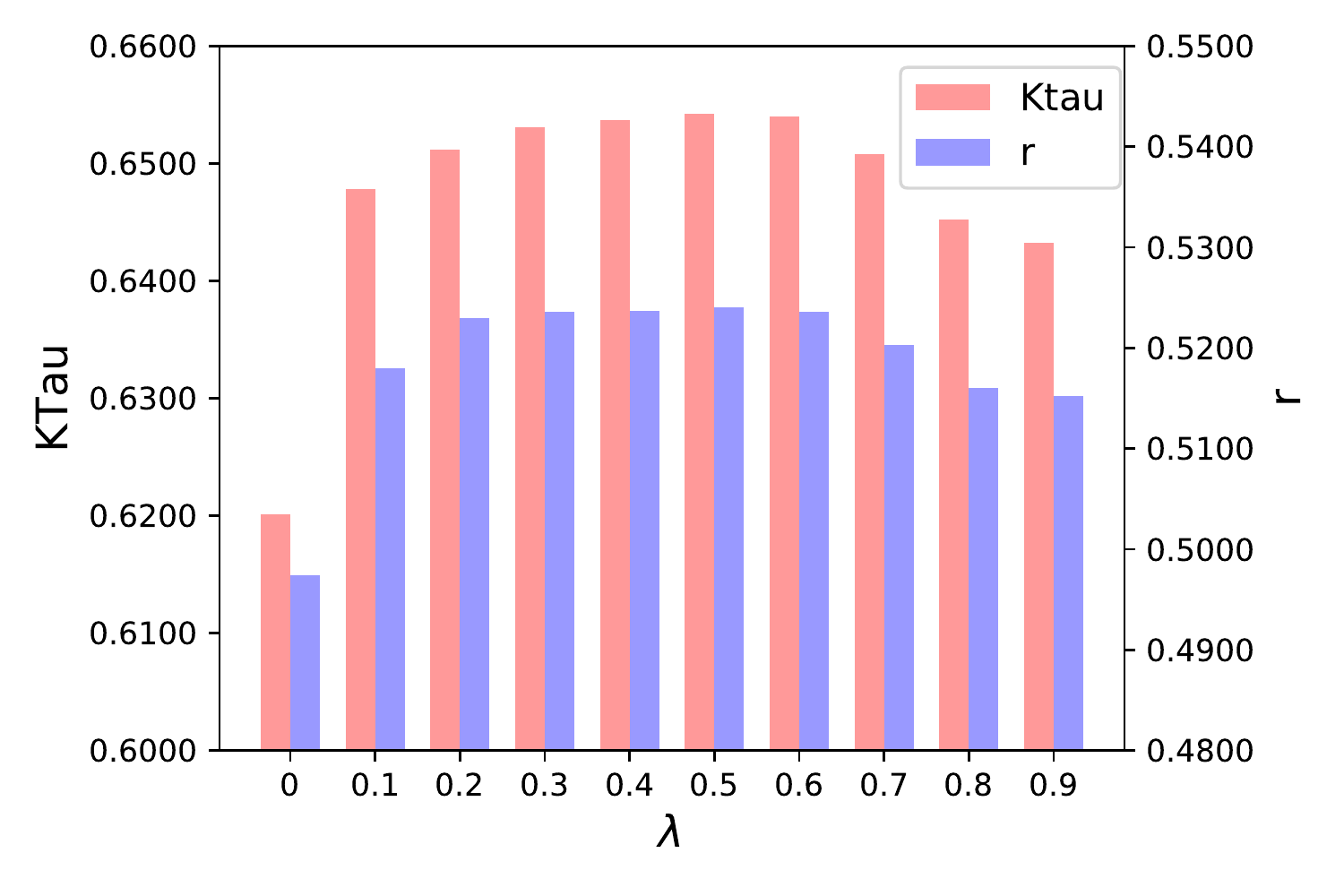}
		\end{minipage}
	}	
	\subfigure[$N^u$]{
		\begin{minipage}[t]{0.33\linewidth}
			\centering
			\includegraphics[width=0.99\linewidth]{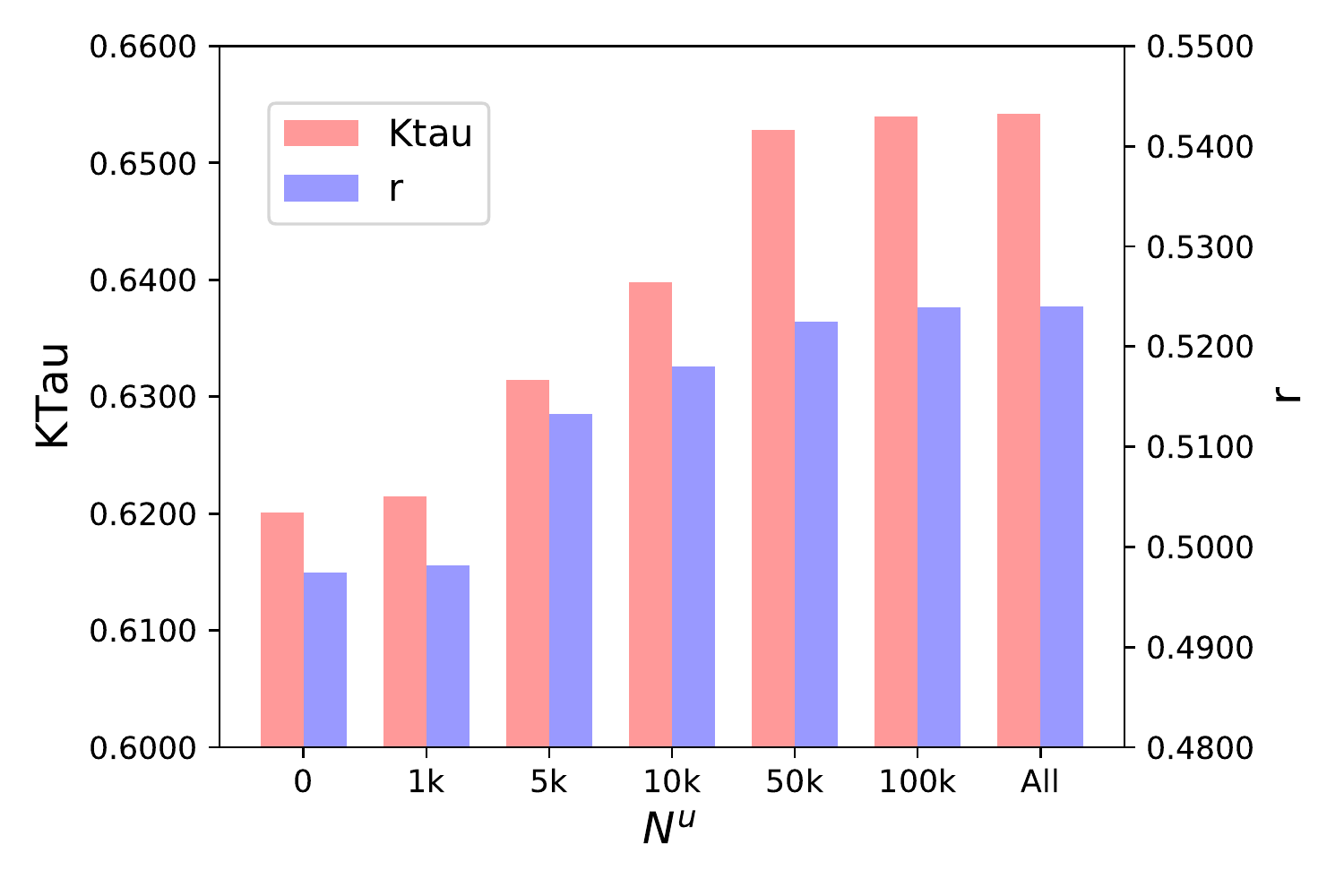}
		\end{minipage}
	}			
	\caption{Performance prediction results of the proposed semi-supervised assessor \wrt different scale factor $\sigma$, weight $\lambda$, and the number of unlabeled architectures $N^u$.}
	\label{fig-abs}	
\end{figure*}

\subsection{Ablation study}
\textbf{The impact of scale factor $\sigma$.} The hyper-parameter $\sigma$ impacts the similarity measurement in Eq.~(\ref{similarity}) and thereby impacts the construct of the graph. With a fixed threshold $\tau$, a denser graph $G$ is constructed with a bigger $\sigma$, and more interaction between different architectures is applied when predicting performance with the GCN assessor. The prediction results with different scale factor $\sigma$ are shown in Figure~\ref{fig-abs}(a), which verifies the effectiveness of utilizing unlabeled architectures with a relation graph to train a more accurate performance predictor. An excessive $\sigma$ also incurs the drop of accuracies in Figure~\ref{fig-abs}(a), as putting too much attention on other architectures also disturb the supervision training process.

\textbf{The impact of weight $\lambda$.} The weight $\lambda$ balances the regression loss $\L_{rg}$ and the reconstruction loss $\L_{rc}$. When the reconstruction loss do not participate in the training process ($\lambda=0$),  prediction accuracies (Ktau and $r$) are lower than those with reconstruction loss as shown in Figure~\ref{fig-abs}(b), since the information in massive unlabeled architectures is not well preserved when constructing the learned architecture representation. 

\textbf{The number of unlabeled architectures $N^u$.} The unlabeled architectures can provide extra information to assist the training of the architecture assessor to make an accurate prediction. As shown in Figure \ref{fig-abs}, with the increasing of unlabeled architectures, both the two criteria KTau and $r$ are increased correspondingly, indicating more accurate performance prediction. The improvement of accuracies comes from that more information is provided by the unlabeled architectures. When the number of unlabeled architectures is enough to reflect the property of the search space (\eg, $N^u=50k$), adding extra unlabeled architectures only brings limited accuracy improvement.

\begin{table}[t]
	\caption{Comparison of prediction accuracies (Ktau) with or without the auto-encoder on Nas-Bench-101 dataset.}
	\label{tb-woencoder}
	\centering
	\begin{tabular}{c||c|c}	\hline
		$N_l$ &W/o Auto-encoder& Ours\\ \hline \hline	
		1k&$0.5302_{\pm0.0081}$&${0.6541}_{\pm0.0078}$\\	\hline
		10k&$0.7188_{\pm0.0025}$&${0.7814}_{\pm0.0042}$\\ \hline	
		100k&$0.7578_{\pm0.0038}$&${0.8456}_{\pm0.0031}$\\ \hline	 			
	\end{tabular}
\end{table}

\textbf{The effect of auto-encoder.}
To show the superiority of the learned representations compared with the hand-craft representations, the prediction results with or without the auto-encoder are shown in Table~\ref{tb-woencoder}. The prediction accuracies (Ktau) are improved obviously by the deep auto-encoder (\eg,  0.6541 \textit{v.s.} 0.5302 with 1k labeled architectures), which indicates the learned representations can reflect the intrinsic characteristics and more compatible to measure architecture similarity and used as the inputs of performance predictor.

\section{Conclusion}
The paper proposes a semi-supervised assessor to evaluate the network architectures by predicting their performance directly. Different from the conventional performance predictors trained in a fully supervised way, the proposed semi-supervised assessor takes advantage of the massive unlabeled architectures in the search space by exploring the intrinsic similarity between architectures. Meaningful representations of architectures are discovered by an auto-encoder and a relation graph involving both labeled and unlabeled architectures is constructed based on the learned representations. The GCN assessor takes both the representations and relation graph to predict the performance. With only 1k architectures randomly sampled from the large NAS-Benchmark-101 dataset \cite{ying2019bench}, the architecture with $94.01\%$ accuracy (top $0.01\%$ of the entire search space) can be found with the proposed method. We plan to investigate the sampling strategy to construct more representative training sets for the assessor and identify better architectures with fewer labeled architectures in the future.

\section*{ Acknowledgment}
This work is supported by National Natural Science Foundation of China under Grant No. 61876007, 61872012, National
Key R\&D Program of China (2019YFF0302902), Australian Research Council under Project DE-180101438, and Beijing Academy of Artificial Intelligence (BAAI).
{\small
	\bibliographystyle{ieee_fullname}
	\bibliography{egbib}
}

\end{document}